# Invariant feature extraction from event based stimuli

Thusitha N. Chandrapala[1] and Bertram E. Shi[2]

*Abstract*—We propose a novel architecture, the event-based Generative Adaptive Subspace Self-Organizing Map (GASSOM) for learning and extracting invariant representations from event streams originating from neuromorphic vision sensors. The framework is inspired by feed-forward cortical models for visual processing. The model, which is based on the concepts of sparsity and temporal slowness, is able to learn feature extractors that resemble neurons in the primary visual cortex. Layers of units in the proposed model can be cascaded to learn feature extractors with different levels of complexity and selectivity. We explore the applicability of the framework on real world tasks by using the learned network for object recognition. The proposed model achieve higher classification accuracy compared to other state-of-the-art event based processing methods. Our results also demonstrate the generality and robustness of the method, as the recognizers for different data sets and different tasks all used the same set of learned feature detectors, which were trained on data collected independently of the testing data.

## I. INTRODUCTION

Event based neuromorphic vision sensors [1-4] have emerged as a viable replacement to frame based sensors in certain applications. These sensors consist of an array of pixels that detect temporal changes in illumination at that spatial location. Increases and decreases in illumination are detected as on and off events. An Address Event Representation (AER) [5] scheme encodes the events by the pixel address and the event type. The advantages of the event based sensor over its traditional frame based counterpart includes low power consumption, low data rate, high dynamic range and high temporal precision. However, processing event based data streams is challenging, because of the discrete and highly variable nature of the data.

An obvious way to process event based data is to accumulate events over fixed time intervals to recreate frames, and then extract information from the recreated frames using standard algorithms developed for frame based cameras. This naïve approach is inefficient as it ignores key advantages of event based sensors, in particular low data rate and high temporal precession. Directly using asynchronous mechanisms for processing event based data will be more cost efficient. For example, Spiking Neural Network architectures (SNNs) inspired by the brain have been identified as a tool that could be utilized in this task. The applicability of simple hierarchical SNN models for visual processing are discussed in [6, 7]. The applicability of hierarchical SNNs for object recognition has been discussed in [8], [9] and [10]. Learning in these networks was limited to the classification network. The spatial structure of the receptive fields of the feature detectors was fixed.

This article focuses on learning invariant feature extractors from event based stimuli, and studies their applicability in object recognition tasks. The events generated by the sensor are dependent on the relative motion between the sensor and the object. Feature extractors applicable for such stimuli should be invariant to the fast changes in the input signal, and extract the underlying static object attributes. In [11] a hierarchical feature extraction method based on learning time-surface prototypes has been introduced, but the issue of invariance was not addressed explicitly.

We present a multilayer architecture that can learn feature extractors from event based data, based on maximizing sparsity and temporal slowness. The concept of temporal slowness ensures that neurons encode the slowly varying component of the signal [12]. It has been previously shown that integrating the concept of temporal slowness has a positive impact in learning invariant feature extractors for frame based inputs [13]. The model consists of layers of processing units that can be stacked such that increasingly abstract features are extracted at higher levels. The feature extractors in each layer are learned using the same generic algorithm. We show that when the network is trained with a stream of events, feature extractors that resemble Gabor filters emerge in the first layer. The units in the next layer learn to encode more complex features, such as corners and extended edges. We test the applicability of the learned feature extractors for object recognition, and obtain better classification accuracy than other state-of-the-art models.

## II. CAMERA SETUP AND DATA ACQUISITION

Event based neuromorphic sensors capture only illumination changes at each pixel. If the scene illumination is constant, the illumination changes observed by the sensor are generated by the interplay between spatial image gradients and the image motion. Relative motion between the sensor and the object is required to generate events. For static environments, the sensor itself must be in motion. Biological systems may exploit a similar strategy. It has been suggested that fixational eye movements play a fundamental role in feature extraction by removing spatial correlations [14] [15].

In order to generate sensor motion, we employ a fixation eye movement model based on [15]. The Dynamic Vision Sensor (DVS) is mounted on a pan-tilt head that is actuated by two independent servo motors. We generate random drift movements which are analogous to the drift movements of the human eye. The drift is modeled as a discrete diffusion process which resets to the center when a boundary is hit. The boundary is defined to be 30 degrees from the center location.

T.N.Chandrapala (tnc@ust.hk) and Bertram E. Shi (eebert@ust.hk) are with the Hong Kong University of Science and Technology.

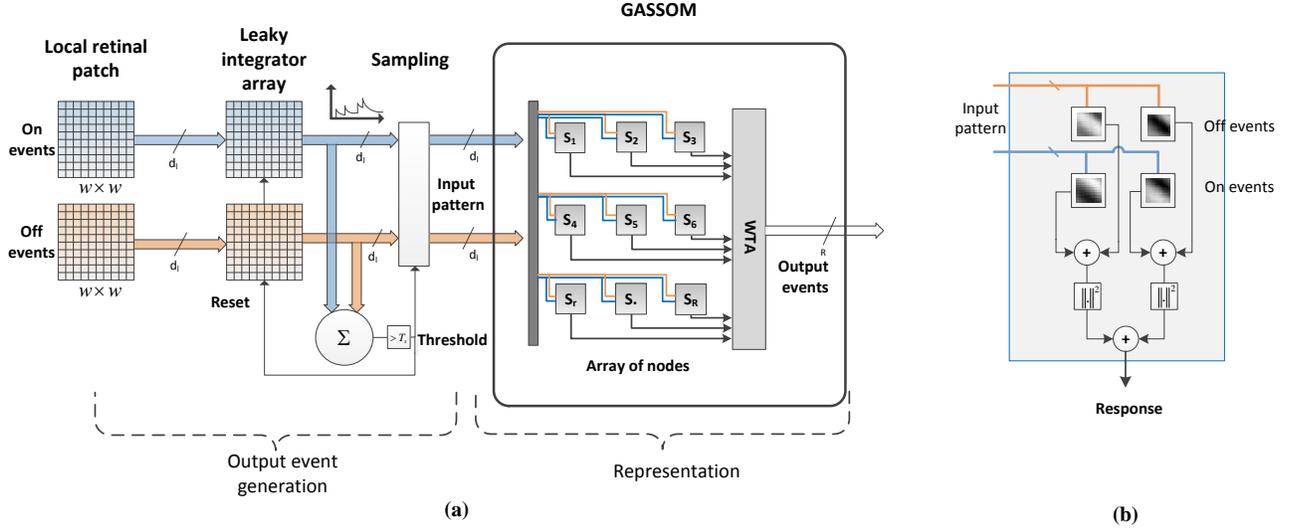

Figure 1. a) The proposed event based processing mechanism. The sub-regions are processed independently. The output event generation is based on a leaky integrate and fire mechanism. When an output event occurs, the network picks a specific node to represent the pattern based on a competitive process. The nodes are updated online to better reflect the environment. (b) The structure of a single subspace for inputs with on and off events.

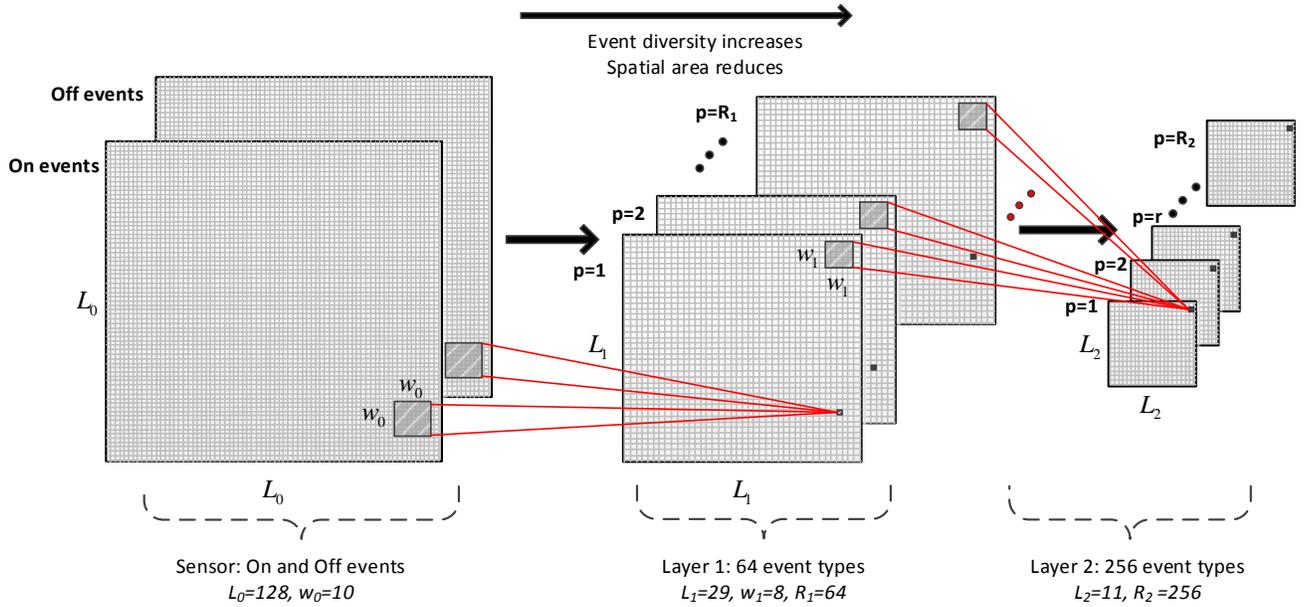

Figure 2. Hierarchical processing of events. Subsequent layers process a smaller spatial area but have larger event diversities.

A diffusion constant of 40 arcmin$^2$s$^{-1}$ is used to generate the drift motion with a discrete time interval of 25ms. While performing these drift movements, the camera is directed at the center of a computer screen where static images are presented with a 2 second time interval. The continuous drift motion along with the presentation of different images is analogous to the situation where the human eye is observing a static environment. Changes in the image on the monitor is analogous to saccades. The drift models fixational eye movements. The images are taken from the Amsterdam Library of Object Images (ALOI) [16]. Since the objects had predominantly vertical and horizontal edges, in order to introduce more diagonal edges, 30% of images were randomly selected to be rotated by a random angle sampled from a uniform distribution between 0 and 180 degrees.

III. SYSTEM ARCHITECTURE

The outline of a single layer of the proposed architecture for feature extraction from event streams is given in Figure 1. The total pixel space of the sensor is broken into $K$ independent $w \times w$ sub-regions indexed by $k$. The stride between two neighboring sub-regions is $\eta$ pixels. Each sub-region is processed independently in an identical manner. An input event from the sensor may appear in multiple sub-regions.

In the following discussion, we describe in the processing of a single sub-region. In the first step, events from a particular spatial location and with a particular event type are passed through a leaky integrator. Each sub-region generates an event when its total activity exceeds a threshold. At that point, the state of the leaky integrators is passed to a representation mechanism, which determines the event type.

*A. Output event generation*

The $i^{th}$ event in a sub-region occurring at the spatial location $\mathbf{y}_i$ at time $t_i$ with event type $p_i \in [1,\ldots,P]$ is represented by $e_i = [\mathbf{y}_i, t_i, p_i]$. For the first layer, there are two types of events (on and off), so $P = 2$. For each sub-region, we construct a spatiotemporal surface by applying leaky integrators to the events coming from a particular location and with a particular type. This is carried out by aggregating the events with an exponential decay with time constant $\tau_f$. The spatiotemporal surface is given by $\mathcal{S}(\mathbf{y}, t, p)$ where the event type is indicated by $p$, the spatial-location by $\mathbf{y}$ and time by $t$. The surface is updated with the arrival of each event from the sub-region. Let the incoming event be given by $e_i = [\mathbf{y}_i, t_i, p_i]$ and the time of the previous event from the sub-region be $t_{i-1}$. The spatiotemporal surface is updated according to

$$\mathcal{S}(\mathbf{y}, t_i, p) = \exp\left(-\frac{(t_i - t_{i-1})}{\tau_f}\right) \mathcal{S}(\mathbf{y}, t_{i-1}, p) + \delta(\mathbf{y} - \mathbf{y}_i, p - p_i) \quad (1)$$

where $\delta(\mathbf{y}, p)$ is the discrete Kroneker delta function which is 1 when the arguments are zero and 0 otherwise. After each input event, the total activity level is calculated according to

$$a(t_i) = \sum_{\forall \mathbf{y}, \forall p} \mathcal{S}(\mathbf{y}, t_i, p) \quad (2)$$

The sub-region generates an output event if the value of $a(t_i)$ is larger than a spiking threshold $T_s$. If an output event is generated, then the value spatiotemporal surface at that time instant is sent to the representation mechanism described below, which determines the event type of the output event. After sampling, all integrators in the sub-region are reset to zero.

*B. Representation mechanism*

We denote the time of the $j^{th}$ output event from the sub-region as $t_j$, to emphasize that not every input event $i$ results in an output event $j$. When an event is generated, we first convert the values of the spatiotemporal surface $\mathcal{S}(\mathbf{y}, t_j, p)$ to a single column vector $\mathbf{x}_j \in \mathcal{R}^D$ where $D = w^2 P$. The mean value of this vector is subtracted from each element, so that it sums to zero, and the vector is normalized so that it has unit norm.

A representation for the signal is produced by a modified version of the online Generative Adaptive Self Organized Map (GASSOM) [13, 17]. The GASSOM learns a set of low dimensional subspaces to encode stimuli using a generative formulation based on Hidden Markov Models [18]. A GASSOM consists of a set of $R$ subspaces indexed by $r \in \{1\ldots R\}$. Each subspace corresponds to a node, with nodes arranged in a lattice in a 2-dimensional latent space. The subspaces are defined by a set of matrices $\mathbf{B}_r = [\mathbf{b}_{r1} \ldots \mathbf{b}_{rh} \ldots \mathbf{b}_{rH}] \in \mathbf{R}^{D \times H}$, where the columns of each matrix represent the basis vectors which span the subspace. Here, we set the subspace dimensionality $H$ to be 2.

Each subspace corresponds to an output event type. The output event type for the $j^{th}$ event is obtained by finding the subspace onto which the vector $\mathbf{x}_j$ has the largest projection. As illustrated in Figure 1 (b), we define the response of each node to the input as the squared length of the projection of an input vector $\mathbf{x}_j$ onto the subspace,

$$\| \mathbf{B}_r^T \mathbf{x}_j \|^2 = \sum_{h=1}^{H} (\mathbf{b}_{rh}^T \mathbf{x}_j)^2 \quad (3)$$

This calculation is similar to that used in energy models of the visual cortical complex cell responses. Basis vectors are analogous to linear simple cell receptive fields, with on and off events processed separately. The output event type $p_j$ is the index of the subspace with largest response.

In order to learn the subspaces online, we assume a generative model for the data, where each data point is generated by one of the nodes, which is identified by the indicator vector $\mathbf{z}(t) \in \{0,1\}^R$ according to a 1 of $R$ coding,

$$\mathbf{z}_j = [z_{j,1} \ldots z_{j,r} \ldots z_{j,R}]^T \quad (4)$$

where,

$$z_{j,r} = \begin{cases} 1 & \text{if } \mathbf{x}_j \text{ is generated by subspace } r \\ 0 & \text{otherwise} \end{cases} \quad (5)$$

Assuming node $r$ generates $\mathbf{x}_j$,

$$\mathbf{x}_j = \mathbf{B}_r \mathbf{w}_{j,r} + \boldsymbol{\varepsilon}_{j,r} \quad (6)$$

where $\mathbf{w}_{j,r} \in \mathbf{R}^H$ is a zero mean Gaussian with a diagonal covariance matrix. The noise $\boldsymbol{\varepsilon}_{j,r}$ is assumed to be orthogonal to the subspace. The nodes generating the data form a Markov chain across time. We assume initially all nodes are equally probable to generate the input. For subsequent times the nodes change based on a transition probabilities that incorporate the concept of temporal slowness. The principle of temporal slowness states that even though the sensory signals vary quickly, the underlying state of the environment which gives rise to the signal varies slowly. The representation evolves to encode the underlying slowly varying signal content.

The transition probability can be modeled as a mix of a uniform distribution over all nodes and a delta function, which controls the probability of self-transition. We control the mixing using a mixing coefficient $0 \leq \rho \leq 1$. The transition probability from node $c$ to $d$ is denoted by $a_{cd}$.

$$a_{cd} = \frac{\rho}{R} + (1-\rho)\delta(c-d) \quad (7)$$

The value of $\rho$ is determined based on the time from last output event using an exponential decay function with time constant $\tau_s$.

$$\rho = 1 - \exp\left(-\frac{t_j - t_{j-1}}{\tau_s}\right) \quad (8)$$

where $t_j$ and $t_{j-1}$ represents time of the current and previous output event respectively. Thus the self-transition probability is high if $t_j - t_{j-1}$ is low.

The learning step first infers the node that generated each input and then updates the corresponding subspaces in the direction of the input. This is done in a similar manner as discussed in [13, 17]. The update to each subspace in the network is given by,

$$\Delta \mathbf{B}_{j,r} = h_{j,r} \cdot \tilde{\mathbf{x}}_{j,r} \cdot \frac{\mathbf{x}_j^T \mathbf{B}_r}{\|\hat{\mathbf{x}}_{j,r}\| \|\mathbf{x}_j\|} \quad (9)$$

where $\hat{\mathbf{x}}_{j,r}$ is the projection of the input onto the subspace $r$,

$$\hat{\mathbf{x}}_{j,r} = \mathbf{B}_r \mathbf{B}_r^T \mathbf{x}_j \quad (10)$$

and $\tilde{\mathbf{x}}_{j,r} = \mathbf{x}_j - \hat{\mathbf{x}}_{j,r}$ is the projection error. The vector of coefficients $\mathbf{h}_j = \begin{bmatrix} h_{j,1} & \cdots & h_{j,R} \end{bmatrix}^T$ is obtained using

$$\mathbf{h}_j = \mathbf{G} \cdot \text{WTA}(\boldsymbol{\gamma}_j) \quad (11)$$

where WTA is the winner take it all function and $\boldsymbol{\gamma}_j = [\gamma_{j,1} \cdots \gamma_{j,R}]^T$ contains the set of conditional probabilities that each node generated the observation $\mathbf{x}_j$ given $\mathcal{X}_j$, the set of observations up to time $t_j$.

$$\gamma_{j,r} = P(z_{j,r} = 1 | \mathcal{X}_j) \quad (12)$$

We refer to $\gamma_{j,r}$ as the responsibility of node $r$ for generating $\mathbf{x}_j$. The values of $\boldsymbol{\gamma}_j$ are obtained iteratively using the forward algorithm. The matrix $\mathbf{G}$ is a Gaussian smoothing matrix, whose elements are given by

$$g_{mn} = \frac{\mathcal{N}(\mathbf{l}_m | \mathbf{l}_n, \sigma \mathbf{I})}{\sum_{k=1}^{R} \mathcal{N}(\mathbf{l}_k | \mathbf{l}_n, \sigma \mathbf{I})} \quad (13)$$

where $\mathcal{N}(\mathbf{l} | \mathbf{m}, \mathbf{C})$ is an n-dimensional Gaussian density function with mean $\mathbf{m}$ and covariance matrix $\mathbf{C}$. $\mathbf{l}_n$ is the indicator of node $n$ in the 2-dimensional latent space.

After updating the subspaces by (9) using an appropriate learning rate, the basis vectors are orthonormalized. This update scheme results in a competitive learning process where only the winner with the highest responsibility score and its neighbors are updated in the direction of the input.

To encode the $K$ input patches, we assume each patch is generated by an independent GASSSOM. As done with convolutional neural networks, we reduce the number of free parameters by assuming that input statistics over the whole visual field are similar, and use the same set of subspaces for each GASSOM.

### C. Cascading feature extractors

The outputs from a single local sub-region map to a single spatial location in the output event space, but could be of $R$ different event types. Therefore output events from the first layer has the same form as the input events, each event corresponding to a particular spatial location, a specific event type, occurring at a specific time. The same mechanism used to process events from the sensor could be used to further process the output events from the first layer, forming a cascaded network. The spatial range of the output events shrink, but based on the number of nodes in the representation network, the specificity of the events increase. The cascaded architecture is shown in Figure 2.

## IV. RESULTS: LEARNED FEATURE EXTRACTORS

The event based framework is implemented in software using C++ and MATLAB (The MathWorks). The first layer is trained using the event stream from the sensor using the ALOI database and the fixation eye movement model described in Section II. After training, the subspaces are fixed and the input is streamed in again to obtain output events from the first layer, which are used to train the second layer. The two layers are trained sequentially to simplify computation. However they could be trained in parallel. The parameters used in training the networks are given in TABLE I.

TABLE I. TRAINING PARAMETERS

| Parameter | Layers | |
|---|---|---|
| | *Layer 1* | *Layer 2* |
| Number of subspaces | 64 | 256 |
| Input event types | 2 | 64 |
| Spatial patch size | 10x10 | 8x8 |
| Stride between patches | 4 | 2 |
| Fire threshold | 40 | 10 |
| $\tau_f$ | 10ms | 50ms |
| $\tau_s$ | 100ms | 50ms |

### A. First layer representation

Each subspace in the GASSOM consists of two basis vectors orthogonal to each other. The first layer basis vectors after training are given in Figure 3. Each sub-image consists of four segments where the left and right columns plot the two basis vectors, and top and bottom parts correspond to on and off events. The basis vectors are similar in structure to Gabor functions with orientation and frequency selectivity. The majority of subspaces, similar to the one highlighted in red, are selective for either on or off events. But we can observe few subspaces, similar to the one highlighted in blue, that are tuned for both on and off events. The top and bottom parts of the basis vectors are phase shifted with respect to each other. This subspace models inputs generated by a moving bar, where one side of the bar generates events of a certain type and the other side generates events of the opposite type.

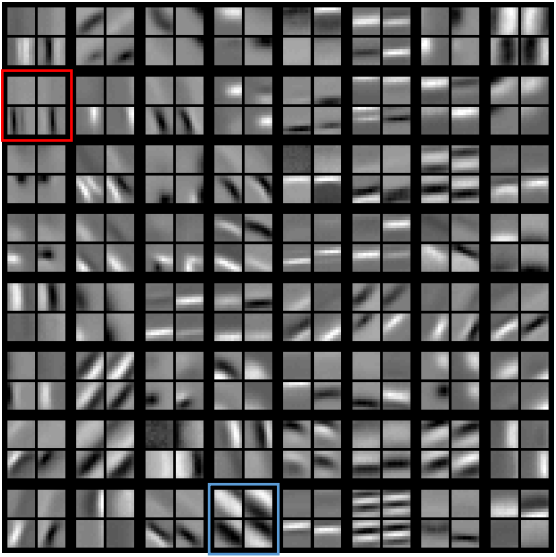

Figure 3. The 64 subspaces learned in the first layer arranged in an 8x8 lattice. Each sub-image corresponds to a subspace. The left and right columns of a sub-image show the two basis vectors which span the subspace. The Top and bottom parts visualize the on and off events. The gray level indicates the weights where white indicates higher than average and black indicates lower than average values.

### B. Second layer representation

The second layer encodes the activation patterns of first layer units. The basis vectors have dimension of $w_1 \times w_1 \times R_1$ ($w_1 = 8$, $R_1 = 64$). We show the structure of second layer subspaces using the corresponding first layer subspaces. A basis vector could be arranged based on the spatial dimensions and the input event type as shown in Figure 4. (a). The value $a(\xi,\psi,p)$ represents the weight attributed to inputs from the spatial location $(\xi,\psi)$ and event type $p$. For each of the $w_1 \times w_1$ spatial locations we obtain the weight with the largest absolute value for all the input event types. This indicates the type of first layer event that would elicit the maximum response from this subspace at that particular spatial location. We plot the selected first layer subspaces for all the corresponding $w_1 \times w_1$ spatial locations and scale the contrast of the individual sub-images according to the absolute value of the associated weight. In Figure 4. (b) we plot a selected set of second layer basis vectors. The patterns of weights suggest that they respond to corners and other higher level structures which are present in images.

### V. RESULTS: OBJECT CLASSIFICATION

For all the classification tasks described in this section, we use the network trained using images from the ALOI database as described in Section III. Note that the images used to train the feature detectors are disjoint from those used in the object classification tasks, and that we use the same feature detectors in all classification tasks.

We use three event based data sets to benchmark the performance of the learned feature detectors. We present the input event streams corresponding to different objects from the benchmark datasets to the network, and record the output events. We discard the spatial and temporal information of the output events, and bin them based on the event type to create a histogram. The normalized histogram is used as the feature for object recognition [11]. Such histograms are calculated for all the instances in the training and testing data. We use the nearest neighbor classification method. The distances between histograms are calculated using the inverse of the intersection of the two histograms [19].

The "Letters and Digits" dataset [8] contains two presentations of 36 characters. The characters are printed on a barrel which rotates at 40rpm in front of an event based camera with a sensor size of 128x128 pixels. Since only two presentations of each character is available, we simply divide the set into two, and use them as the training and testing datasets. We were able to achieve 100% classification accuracy using the first layer of the trained network.

The "Faces" data set introduced in [11] poses a more difficult classification task. This dataset contains 24 samples each from 7 subjects. The event stream is obtained using the ATIS camera [1] with a sensor of 320x240 pixels. The sensor is kept stationary while the subjects move their heads to trace a square path generated on a computer monitor. We obtain 100% accuracy for this classification task using a single layer of the network. We use eight-fold cross validation to average the result.

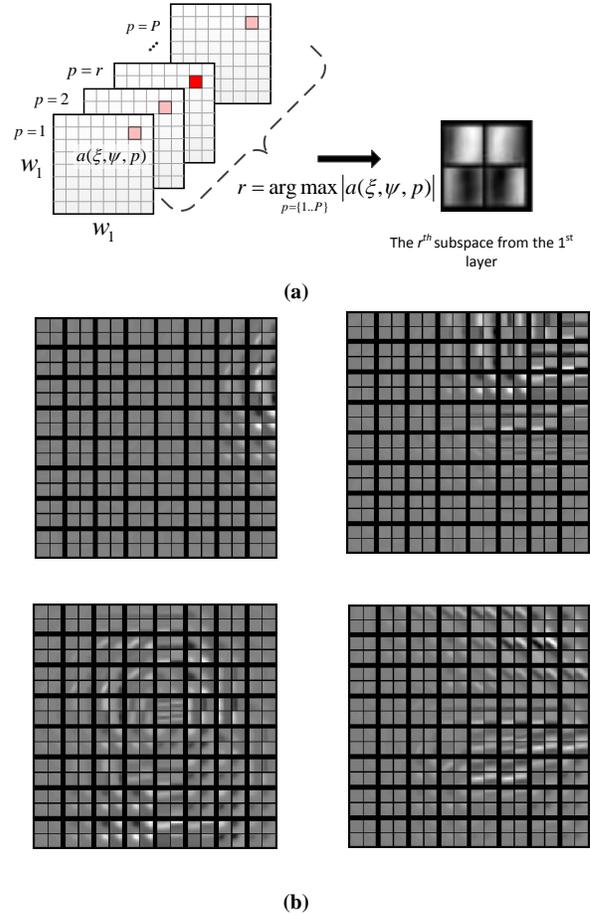

Figure 4. Visualization of selected 2nd layer basis vectors. using the 1st layer basis vectors. (a) the 3-D representation of a 2nd layer basis vector, (b) The 2-D visualization of 2nd layer basis vectors by selecting 1st layer units.

The "Event based MNIST" database [20] proved to be the most difficult task because of the variability of the original hand written images [21]. The data set contains 60,000 training instances and 10,000 testing instances. The original dataset has been converted to the event domain by displaying each digit on a computer screen and having the camera make 3 quick motions tracing an isosceles triangle. Using the first layer of the network, we achieve a classification accuracy of 83%. With the combined first and second layer features, we achieve an accuracy of 89.9%. A higher classification accuracy (95.72%) for the event based MNIST database has been reported in [22], but in this work the network has been trained on frame based data and later converted to event domain.

TABLE II. SUMMARY OF CLASSIFICATION PERFORMANCE

| Dataset | Classification accuracy | |
|---|---|---|
| | Event based GASSOM | Previous best reported |
| Letters and Digits | 100% | 100% [11] |
| Faces | 100% | 79% [11] |
| Event based MNIST | 89.9% | 83.4% [20] |

## VI. Discussion and conclusion

The event based GASSOM presented in this work learns invariant feature extractors from event based stimuli generated from neuromorphic sensors. Learned subspaces resemble the spatial structure of visual cortical cells. We demonstrate high accuracy for three object recognition tasks using representations learned on a completely different set of data. Thus the representations learned by the model are generalizable over event streams originating from different sensors and conditions.

We use a simple histogram distance based nearest neighbor classifier, and discard temporal information. Even though the three datasets used in object classification have deterministic motion which is repeated over all instances, if a dataset is created using pseudo-random drift like motion, the order of output events become unreliable. Therefore in this work we focus only on spatial patterns like edges and corners for object identification.

## Acknowledgment

We offer our gratitude to Jörg Conrad of Technische Universität München for providing us with the DVS camera setup and related advice.

## References

[1] C. Posch, D. Matolin and R. Wohlgenannt, "A QVGA 143 dB dynamic range frame-free PWM image sensor with lossless pixel-level video compression and time-domain CDS," *IEEE Journal of Solid-State Circuits*, pp. 259-275, 2011.
[2] T. Serrano-Gotarredona and B. Linares-Barranco, "A 128 128 1.5% Contrast Sensitivity 0.9% FPN 3 μs Latency 4 mW Asynchronous Frame-Free Dynamic Vision Sensor Using Transimpedance Preamplifiers," *IEEE Journal Of Solid-State Circuits*, vol. 48, pp. 827-838, 2013.
[3] P. Lichtsteiner, C. Posch and T. Delbruck, "A 128× 128 120 dB 15 μs latency asynchronous temporal contrast vision sensor," *IEEE Journal Of Solid-State Circuits*, vol. 43, pp. 566-576, 2008.
[4] M. A. Mahowald and C. Mead, "The Silicon Retina," *Scientific American*, 1991.
[5] K. Boahen, "Point-to-point connectivity between neuromorphic chips using address events," *IEEE Transactions on Circuits and Systems II: Analog and Digital Signal Processing*, vol. 47, pp. 416-434, 2000.
[6] F. Folowosele, R. J. Vogelstein and R. Etienne-Cummings, "Real-time silicon implementation of V1 in hierarchical visual information processing", *IEEE Biomedical Circuits and Systems Conference*, 2008
[7] R. J. Vogelstein, U. Mallik, E. Culurciello, G. Cauwenberghs and R. Etienne-Cummings, "A multichip neuromorphic system for spike-based visual information processing," *Neural Computation*, vol. 19, pp. 2281-2300, 2007.
[8] G. Orchard, C. Meyer, R. Etienne-Cummings, C. Posch, N. Thakor and R. Benosman, "HFirst: A Temporal Approach to Object Recognition", *IEEE Transactions on Pattern Analysis and Machine Intelligence*, 37(10), pp.2028-2040, 2015.
[9] J. A. Pérez-Carrasco, B. Zhao, C. Serrano, B. Acha, T. Serrano-Gotarredona, S. Chen and B. Linares-Barranco, "Mapping from Frame-Driven to Frame-Free Event-Driven Vision Systems by Low-Rate Rate Coding and Coincidence Processing Application to Feedforward ConvNets," *IEEE Transactions On Pattern Analysis and Machine Intelligence*, vol. 35, pp. 2706-2719, 2013.
[10] B. Zhao, R. Ding, S. Chen, B. Linares-Barranco and H. Tang, "Feedforward categorization on AER motion events using cortex-like features in a spiking neural networks", *IEEE Transactions on Neural Networks and Learning Systems*, pp.1963-1978, 2014.
[11] X. Lagorce, G. Orchard, F. Gallupi, B. Shi and R. Benosman, "HOTS: A Hierarchy of Event-based Time-Surfaces for pattern recognition," *IEEE Transactions on Pattern Analysis and Machine Intelligence* (to appear).
[12] P. Földiák, "Learning invariance from transformation sequences," *Neural Computation*, vol. 3, pp. 194-200, 1991.
[13] T. N. Chandrapala and B. E. Shi, "Learning Slowness in a Sparse Model of Invariant Feature Detection," *Neural Computation*, 2015.
[14] C. Cherici, X. Kuang, M. Poletti and M. Rucci, "Precision of sustained fixation in trained and untrained observers," *Journal of Vision*, vol. 12, 2012.
[15] X. Kuang, M. Poletti, J. D. Victor and M. Rucci, "Temporal encoding of spatial information during active visual fixation," *Current Biology*, vol. 22, pp. 510-514, 2012.
[16] J. Geusebroek, G. J. Burghouts and A. W. Smeulders, "The Amsterdam library of object images," *International Journal of Computer Vision*, vol. 61, pp. 103-112, 2005.
[17] T. N. Chandrapala and B. E. Shi, "The Generative Adaptive Subspace Self-Organizing Map," in *IEEE International Joint Conference on Neural Networks*, Beijing, 2014.
[18] L. R. Rabiner, "A tutorial on hidden Markov models and selected applications in speech recognition," *Proceedings of the IEEE*, vol. 77, pp. 257-286, 1989.
[19] S. Cha and S. N. Srihari, "On measuring the distance between histograms," *Pattern Recognition*, vol. 35, pp. 1355-1370, 2002.
[20] G. Orchard, A. Jayawant, G. Cohen and N. Thakor, "Converting Static Image Datasets to Spiking Neuromorphic Datasets Using Saccades," *arXiv Preprint*, arXiv:1507.07629, 2015.
[21] Y. LeCun, L. Bottou, Y. Bengio and P. Haffner, "Gradient-based learning applied to document recognition," *Proceedings of the IEEE*, vol. 86, pp. 2278-2324, 1998.
[22] D. Neil and S. Liu, "Effective Sensor Fusion with Event-Based Sensors and Deep Network Architectures," *IEEE International Symposium on Circuits and Systems*, Canada, 2016